\documentclass{article}

\PassOptionsToPackage{square,numbers}{natbib}

\usepackage[final]{nips_2016}

\usepackage[utf8]{inputenc} 
\usepackage[T1]{fontenc}    
\usepackage{url}            
\usepackage{booktabs}       
\usepackage{amsfonts}       
\usepackage{nicefrac}       
\usepackage{microtype}      

\usepackage{subcaption}
\usepackage[ruled]{algorithm2e}

\usepackage{graphicx}
\usepackage[font=small]{caption}

\title{3D Simulation for Robot Arm Control with Deep Q-Learning}

\author{
  Stephen James \\
  Imperial College London\\
  \texttt{slj12@imperial.ac.uk} \\
  \And
  Edward Johns \\
  Imperial College London \\
  \texttt{e.johns@imperial.ac.uk} \\
}

\begin{document}

\maketitle

\begin{abstract}
Recent trends in robot arm control have seen a shift towards end-to-end solutions, using deep reinforcement learning to learn a controller directly from raw sensor data, rather than relying on a hand-crafted, modular pipeline. However, the high dimensionality of the state space often means that it is impractical to generate sufficient training data with real-world experiments. As an alternative solution, we propose to learn a robot controller in simulation, with the potential of then transferring this to a real robot. Building upon the recent success of deep Q-networks, we present an approach which uses 3D simulations to train a 7-DOF robotic arm in a control task without any prior knowledge. The controller accepts images of the environment as its only input, and outputs motor actions for the task of locating and grasping a cube, over a range of initial configurations. To encourage efficient learning, a structured reward function is designed with intermediate rewards. We also present preliminary results in direct transfer of policies over to a real robot, without any further training.
\end{abstract}

\section{Introduction}

Traditionally, robot arm control has been solved by hand-crafting solutions in a modular fashion, for example: 3D reconstruction, scene segmentation, object recognition, object pose estimation, robot pose estimation, and finally trajectory planning. However, this approach can result in loss of information between each of the modules resulting in accumulation of error, and it is not flexible for learning a range of tasks, due to the need for prior knowledge.

A recent trend has seen robot arm control being learned directly from image pixels, in an end-to-end manner using deep reinforcement learning. This exploits the power of deep learning with its ability to learn complex functions from raw data, avoiding the need for hand-engineering cumbersome and complex modules. However, applying deep reinforcement learning to robot control is challenging due to the need for large amounts of training data to deal with the high dimensionality of policies and state space, which makes efficient policy search difficult. 

Deep reinforcement learning for robot manipulation often uses real-world robotic platforms, which usually require human interaction and long training times. We believe that a promising alternative solution is using simulation to train a controller, which can then be directly applied to real-world hardware, scaling up the available training data without the burden of requiring human interaction during the training process.

Building upon the recent success of deep Q-networks (DQNs) \cite{HumanLevelControl}, we present an approach that uses deep Q-learning and 3D simulations to train a 7-DOF robotic arm in a robot control task in an end-to-end manner, without any prior knowledge. Images are rendered of the virtual environment and passed through the network to produce motor actions. To ensure that the agent explores interesting states of this 3D environment, we use intermediate rewards that guide the policy to higher-reward states. As a test case for evaluation, our experiments focus on a task that involves locating a cube, grasping it, and then finally lifting it off a table. Whilst this particular task is simpler than some others trained with real-world data, the novelty of end-to-end deep reinforcement learning requires evaluation of these initial experiments in order to set the scene for further, more complex applications in future work.  

\section{Related work}

Deep reinforcement learning enables policies to be learned by exploiting the function approximation powers of deep neural networks, to map observations directly to actions. Recently, state-of-the-art results have been achieved with deep Q-networks (DQN) \cite{HumanLevelControl}, a novel variant of Q-learning that is used to play 49 Atari-2600 games using only pixels as input. This method was designed for use with discrete action spaces, and is adopted for our work. However, recent extensions adapted deep Q-learning for the continuous domain \cite{ContinuousControlWithDeepReinforcementLearning}, allowing high dimensional action spaces such as that in physical control tasks.
 
For robot manipulation, recent work has also seen deep reinforcement learning used to learn visually-guided robot arm controllers via convolutional neural networks \cite{EndToEndTraining, DeepSpatialAutoencoders, LearningContactRichManipulationSkills}. Most notably, guided policy search \cite{GPS, EndToEndTraining} has accomplished tasks such as block insertion into a shape sorting cube and screwing on a bottle cap, by using trajectory optimization to direct policy learning. While the tasks undertaken in these papers are impressive, they require significant human involvement for data collection. An alternative approach is to use large-scale self-supervision \cite{SupersizingSelfSupervision, LearningHandEyeCoordination, DeepReinforcementLearningForRoboticManipulation, DeepVisualForesightForPlanningRobotMotion}, but in practice this is limited due to the expense of acquiring many robots for learning in parallel, or the time consuming nature if only a single robot is available.

We believe that training in simulation is a more scalable solution for the most complex manipulation tasks. Simulation has been used for a number of tasks in computer vision and robotics, including object recognition \cite{PairwiseDecomposition}, semantic segmentation \cite{UnderstandingScenesWithSynthetic}, robot grasping \cite{DeepLearningAGraspFunction}, and autonomous driving \cite{SynthiaDataset}. For robot arm control, it was recently shown that policies could be learned in simulation for a 2D target reaching task \cite{TowardsVisionBasedDeepReinforcementLearning}, but this failed to show any feasibility of transferring to the real world. To address this issue of transfer learning, a cross-domain loss was proposed in \cite{TowardsAdaptingDeepVisuomotorRepresentations} to incorporate both simulated and real-world data within the same loss function. An alternative approach has made use of progressive neural networks \cite{ProgressiveNeuralNetworks}, which ensure that information from simulation is not forgotten when further training is carried out on a real robot \cite{SimToRealRobotLearning}. However, our approach differs in that we do not require any real-world training, and attempt to directly apply policies learned in simulation, over to a real robot.

\section{Approach}

Our approach uses a DQN \cite{HumanLevelControl} which approximates the optimal action-value (or Q) function for a given input:
\begin{equation}
Q^*(s,a) = \max_{\pi} E_\pi(\sum_{k=0}^{\infty} \gamma^k r_{t+k} | s_t = s, a_t = a).
\label{eq:q_learning}
\end{equation}
Here, $Q(s,a)$ defines the \emph{q-value} of action $a$ and state $s$. A \emph{policy} $\pi$ determines which action to take in each state, and is typically the action associated with the highest q-value for that state. $Q^*(s,a)$ is then the optimum policy, defined as that which has the greatest expected cumulative reward $r$ when this policy is followed, over each time step $k$ between the starting state at time $t$ and some termination criteria (or infinite time). The discount factor $\lambda$ gives weight to short-term rewards and ensures convergence of Equation \ref{eq:q_learning}.

The Q-learning update rule uses the following loss function at iteration $i$:
\begin{equation}
L_i(\theta_i) = E_{(s,a,r,s') \sim \mathcal{D}} \Big[ r + \gamma \max_{a'} Q(s', a'; \theta_i^-) - Q(s, a; \theta_i) \Big]^2,
\end{equation}
where $\theta_i$ are the parameters of the Q-network, and $\theta_i^-$ are the parameters used to compute the target.

In this paper, we use this approach to train a 3D simulation of a 7-DOF robotic arm in a control task. On each iteration of training, an image is taken of the virtual environment and passed through the network to produce motor actions. We define an episode as 1000 iterations or until the task has been completed. The task is considered completed when the arm has grasped the cube and lifted it to a height of 30cm. At the end of each episode, the scene is reset, and a new episode begins. We define the maximum number of iterations to be 1000 so that the agent does not waste time in episodes that cannot be completed successfully --- for example, if the agent knocks or drops the cube out of its grasping range. We use an $epsilon$-$greedy$ exploration approach which decides between a random action or the action corresponding the highest Q-value produced by the DQN. During training, $epsilon$ is annealed over a period of 1 million episodes.

\begin{figure}[h!]
  \centering
  \includegraphics[width=\linewidth]{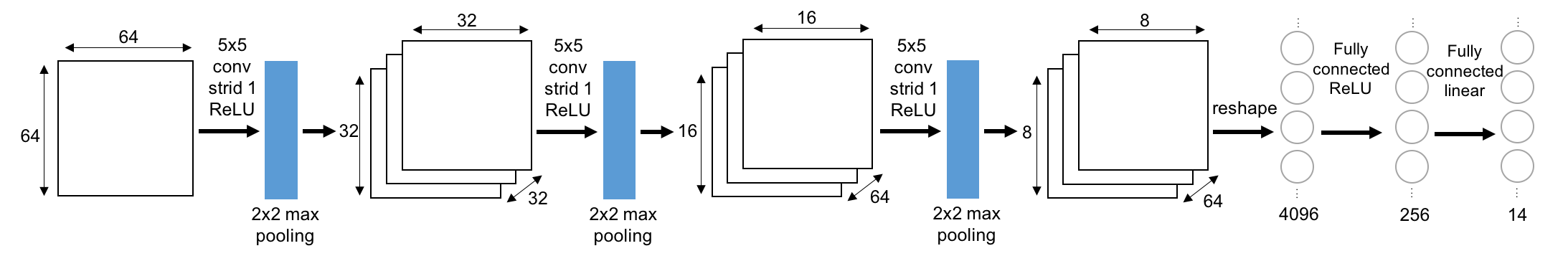}
  \caption{The architecture of our deep Q-network. The network contains 3 convolutional layers with max-pooling layers in-between, followed by 2 fully-connected layers.}
  \label{fig:cnn_architecture}
\end{figure}

Our state is represented by an image of the scene which includes both the arm and cube. We define a set of 14 actions that control the gripper and 6 joints of the arm. Each joint can be controlled by 2 actions --- one action to rotate the joint by one degree clockwise, and another action to rotate one degree counter clockwise. The gripper follows similarly, with one action to open the gripper, and the other to close it. Our network, summarised in Figure \ref{fig:cnn_architecture}, has 14 outputs corresponding to the Q-values for each action given the current state. 

During each of our experiments, we fix the discount rate at $\gamma = 0.99$ and set our replay memory size to $500,000$.  Gradient descent is performed using the Adam optimisation algorithm \cite{Adam} with the learning rate set at $\alpha = 6 \times 10^{-6}$.

Due to the size of the state space, we must engineer the simulator to produce rewards that encourage the agent to explore states around the cube. We use positional information to give the agent intermediate rewards based on the distance between the gripper and the cube. Summarised in Algorithm \ref{alg:reward}, rewards are given to the agent based on the exponential decaying distance from the gripper to the cube ($e^{- \gamma \times distance}$), with a reward of $1$ while the cube is grasped, and then an additional reward based on the distance the cube is from the ground. During our experimentation, we defined $\gamma = 0.25$. When the arm lifts the cube high enough to complete the task, the agent receives a final reward of 100 for task completion.

\begin{figure}[h]
\centering
\begin{minipage}{.325\linewidth}
\begin{algorithm}[H]
\DontPrintSemicolon
\SetAlgoNoLine
\
\uIf{terminal}{
	$r = 100$
}
\uElseIf{target.grasped}{
	$r = 1 + target.position.y$
}
\Else{
	$r = e^{- \gamma \times distance}$
}
\caption{Reward function}
\label{alg:reward}
\end{algorithm}
\end{minipage}
\end{figure}

\section{Experiments and results}

During the course of this research, we used two versions of our simulator. The first version was built from primitive shapes and was used as an initial proof-of-concept to show that policies could be learned when trained and tested in simulation. We then turned our attention to evaluating the feasibility of transferring trained networks to the real world without any further training. In order for policy to have any chance of transferring, we needed to modify the simulation so that the virtual world would resembled the real world as much as possible. This version was then used to train one final network which would be tested in the real world.

\begin{figure}[h]
  \centering
  \includegraphics[width=\linewidth]{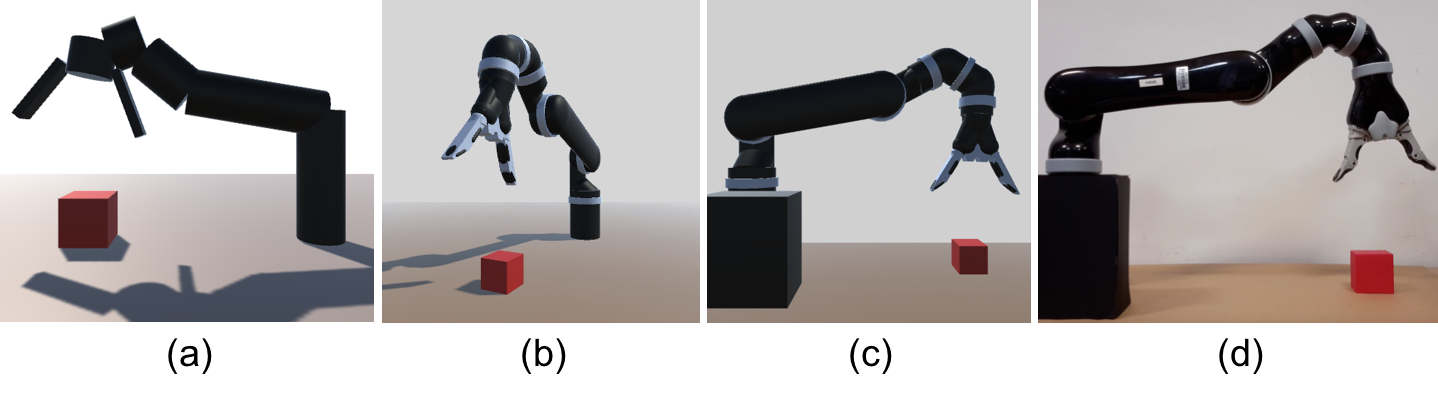}
  \caption{Image (a) shows the first version of our simulation. Both images (b) and (c) show the second version of our simulation that was used to train a policy that would be tested in a real-world environment (image (d)). A black box is placed in front of the base in order to avoid complicated modelling of the clamp holding the arm to the table, and the wires connected to the robot.}
  \label{fig:robot_strip}
\end{figure}

Our first experiment saw the agent learn to grasp and then lift the cube from a fixed starting position shown by image (a) in Figure \ref{fig:robot_strip}. Interestingly, we observed the arm returning to grasp the cube if it was dropped during the lifting motion (due to a random "open gripper" action selected via the $epsilon$-$greedy$ method). This demonstrates the ability of the agent to recognise high rewards associated with grasping the object from a range of states, despite having never been trained for a re-grasping action.

\begin{figure*}
\centering
\begin{subfigure}{0.32\textwidth}
\includegraphics[width=\linewidth]{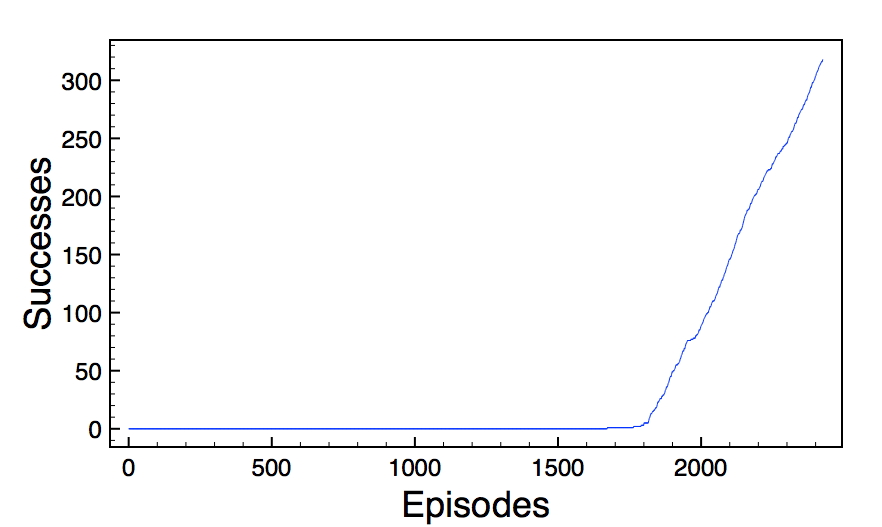} 
\caption{Cumulative successes}
\label{fig:robox_image_no_joints_progress}
\end{subfigure}
\begin{subfigure}{0.32\textwidth}
\includegraphics[width=\linewidth]{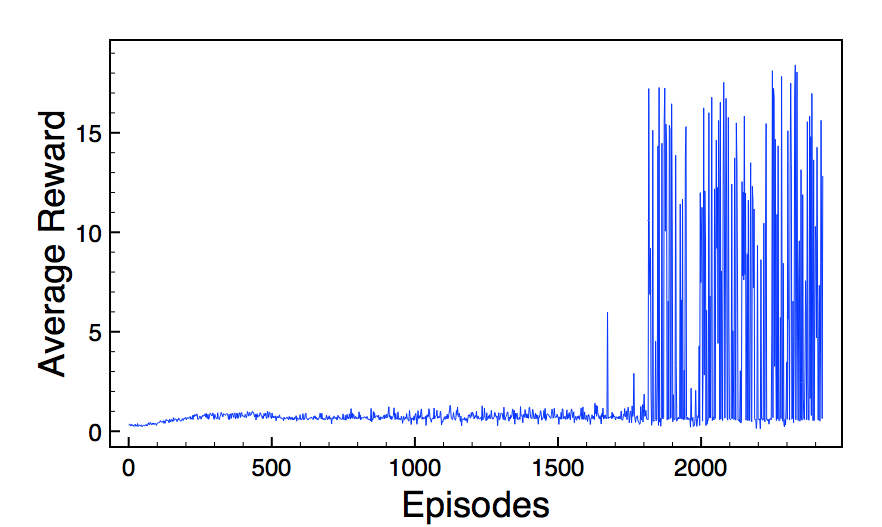}
\caption{Average Reward}
\label{fig:robox_image_no_joints_rewards}
\end{subfigure}
\begin{subfigure}{0.32\textwidth}
\includegraphics[width=\linewidth]{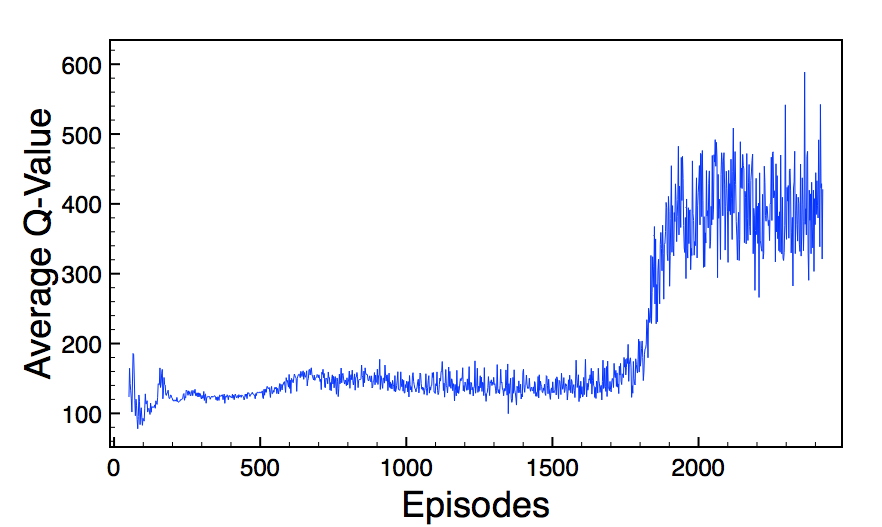}
\caption{Average Q-Value}
\label{fig:robox_image_no_joints_qvalues}
\end{subfigure}
\caption{Experiment with constant cube positions and joint angles using image as input. (a) shows the cumulative successes after each episode, (b) shows the average reward per episode, and (c) shows the average maximal Q-value per episode.}
\label{fig:robox_image_no_joints}
\end{figure*}

In addition to observing the agent in action when being tested to get a good indication of whether the agent has begun to learn a set of high reward policies, we can also study the data collected along the learning process. Figure \ref{fig:robox_image_no_joints} shows us the cumulative successes after each episode, the average reward per episode, and the average maximal Q-value per episode. When the agent starts to succeed on a regular basis at 1800 episodes (Figure \ref{fig:robox_image_no_joints_progress}), we see a greater frequency of increased average rewards (Figure \ref{fig:robox_image_no_joints_rewards}), along with a sharp increase in the average maximal Q-values (Figure \ref{fig:robox_image_no_joints_qvalues}).

Following this, we attempted to make our network learn to deal with a range of starting joint configurations and starting cube positions. At the start of each episode, the joint angles were set in a similar configuration to image (a) in Figure \ref{fig:robot_strip}, but with variations of 20 degrees in each joint. During training, each time the agent was successful in completing the task, the cube was moved to a random location within a $200 cm^2$ rectangular area. A comparison of this expriment and the previous are summarised in Table \ref{results-table}. Here, we see the importance of training for tolerance to varying initial conditions compared to training with a fixed initial condition, where the success rate jumps from 2\% to 52\%.

\begin{table}[h]
  \caption{Success rate when running 50 episodes with $\epsilon = 0.1$. Environment A refers to keeping the joint angles and cube position constant at the start of each episode, while Agent A is the agent that was trained in this environment. Environment B refers to setting the joint angles and cube position randomly at the start of each episode, while Agent B is the agent that was trained in this environment.}
  \label{results-table}
  \centering
  \begin{tabular}{l|ll}
    \toprule
    Environment & Agent A & Agent B \\
    \midrule
    A           & 56\%    & 64\%    \\
    B           & 2\%     & 52\%    \\
    \bottomrule
  \end{tabular}
\end{table}

\begin{figure}[h]
  \centering
  \includegraphics[width=\linewidth]{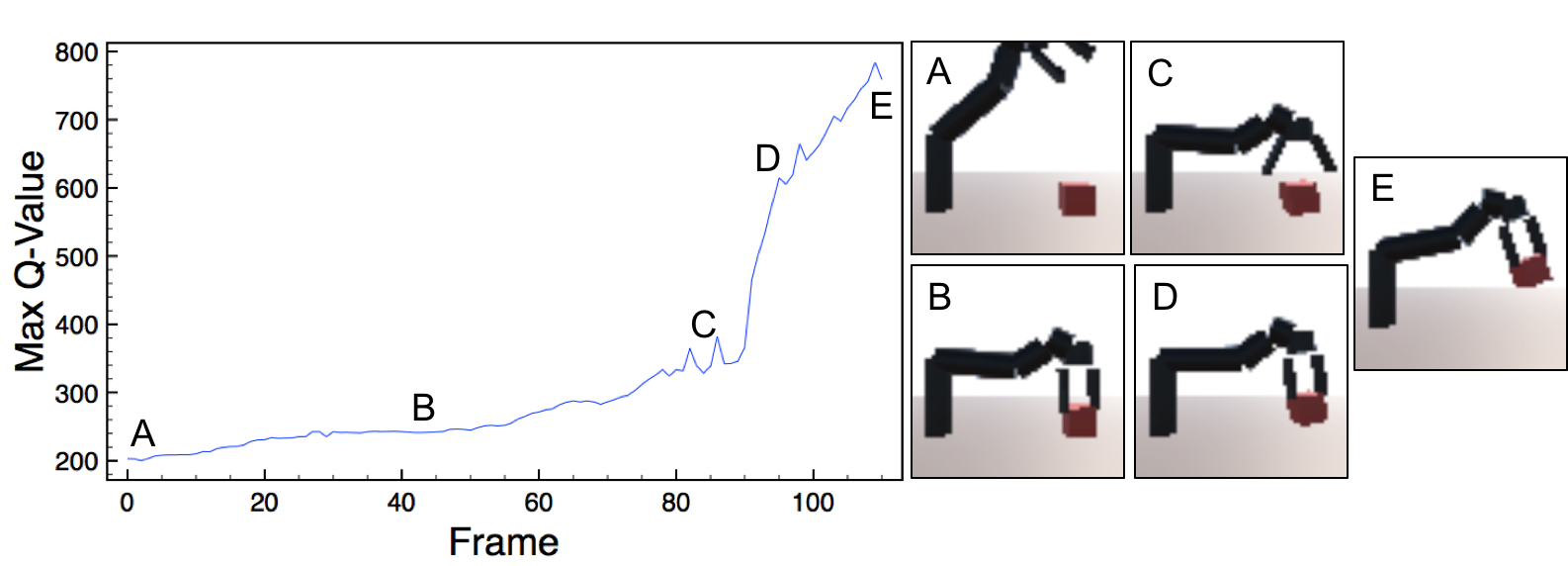}
  \caption{\footnotesize Visualisation of the learned value function on a successful episode of 112 iterations/frames.}
  \label{fig:frame_by_frame}
\end{figure}

Figure \ref{fig:frame_by_frame} shows a visualisation of the learned value function on a successful episode of 112 iterations (or frames). There are 5 images labelled from A to E which correspond to the frame numbers in the graph. The frames from A to C show a steady increase in the Q-values which is a result of the agent getting closer to the cube. At frame C, the Q-values fluctuate briefly due to the network trying to determine whether or not it has grasped the cube. By the time we get to frame D, there has been a large jump in the Q-values where the agent seems certain that it has the cube. The Q-value then continues to rise as the height of the cube increases, and finally peaks as the agent expects to receive a large reward for task completion. The visualisation shows that the value function is able to evolve over time for a robot control task.

To explore direct transfer to a real robot, we first needed to ensure that the simulated world resembled the real-world as much as possible. For this, we set out a scene in the real world and then created a more visually-realistic simulation than the first attempt. Images (c) and (d) in Figure \ref{fig:robot_strip} show both the real-world scene and its simulated replica. We took the network trained on this simulation and ran it directly on the real-world version of the arm with epsilon fixed at $0.1$. As we had hoped, the agent exhibited similar behaviour to its simulated counterpart, moving the gripper directly towards the cube --- suggesting that transferring from simulation to real-world is possible.

However, the action to close the gripper was in fact rarely chosen by our trained policy when applied to the real-world, and so the agent was unable to fully complete the task in one sequence. Instead, the agent typically progresses towards the cube and remains in nearby states. In order to test if the agent could complete the task had it closed its gripper, we started the agent in a position where it has the cube in its grasp. We re-ran the experiment from this starting state, and then the agent completed the task successfully by lifting the cube in an upwards direction. This illustrates that transferring control of the gripper is more challenging than that of the 6 joints on the arm, perhaps due to there being less room for error in a binary "open or close" action, compared to an action involving movement in continuous 3D space.

\begin{figure}[h]
  \centering
  \includegraphics[width=\linewidth]{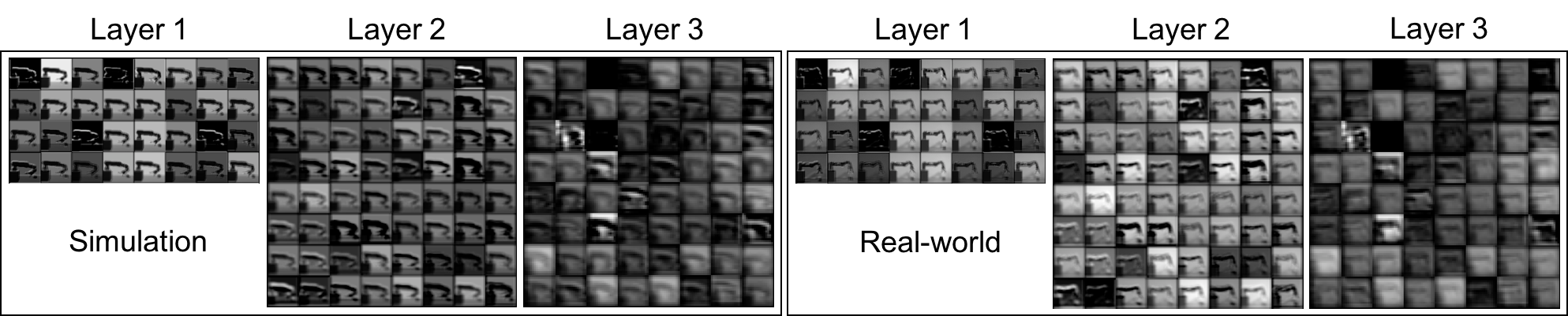}
  \caption{Comparison of the feature map activations of the simulation and real-world. The feature map activations on the left are from simulation while the activations on the right are from the real-world.}
  \label{fig:activations_sim_real}
\end{figure}

Finally, observing the feature map activations in Figure \ref{fig:activations_sim_real} gives evidence that transfer is somewhat successful, by showing similar feature map activations for both simulation and the real world. All layers show very similar activations to the respective counterpart layers.

\section{Conclusions}

The experiments and results in this paper present an investigation into training robot arm control using deep Q-learning via 3D simulation. Whilst the experimental set up is simple in our case, extension to more complex and useful tasks can now be treated as one of scaling up the training data, and increasing the accuracy of graphics and physics simulation. Given the scalability of simulation compared to real-world training as is typically performed elsewhere, the range of possible tasks which could be learned is now significantly greater than before, together with the capacity for generalisation to scenes with new layouts, colours and textures. This paper provides a strong foundation from which highly-scalable, end-to-end training of complex robot arm control, can now be fully investigated.

\end{document}